\documentclass[11pt,a4paper]{article}
\usepackage[hyperref]{emnlp-ijcnlp-2019}
\usepackage{times}
\usepackage{latexsym}
\usepackage{bm}
\usepackage{url}

\usepackage{smile}
\usepackage{latexsym}
\usepackage{longtable}
\usepackage{color,soul}
\usepackage{multirow}

\usepackage{amsmath}
\usepackage{amsfonts,amssymb,amsthm,epsfig,epstopdf,url,array}

\DeclareMathOperator*{\softmax}{softmax}

\DeclareMathOperator*{\sigmoid}{sigmoid}

\graphicspath{{./imgs/}}

\makeatletter
\g@addto@macro\normalsize{%
	\setlength\abovedisplayskip{.2ex}
	\setlength\belowdisplayskip{.15ex}
	\setlength\abovedisplayshortskip{.2ex}
	\setlength\belowdisplayshortskip{.15ex}
}
\makeatother

\aclfinalcopy % Uncomment this line for the final submission

\title{Conditional Self-Attention for Query-based Summarization}

\author{Yujia Xie \thanks{~~Work done during internship in Microsoft. } \\
  Georgia Tech \\
  {\scriptsize \tt Xie.Yujia000@gmail.com} \\\And
  Tianyi Zhou \\
  University of Washington \\
  {\scriptsize \tt tianyi.david.zhou@gmail.com}
  \\\And 
  Yi Mao \\
  Microsoft \\
  {\scriptsize \tt maoyi@microsoft.com}
  \\\And
  Weizhu Chen \\
  Microsoft \\
  {\scriptsize \tt wzchen@microsoft.com} \\} 

\date{}

\begin{document}
\maketitle
\begin{abstract}
%Query-based summarization aims to generate a short abstract of a long text that is able to cover the main information relevant to a given query. 
Self-attention mechanisms have achieved great success on a variety of NLP tasks due to its flexibility of capturing dependency between arbitrary positions in a sequence. 
For problems such as query-based summarization (Qsumm) and knowledge graph reasoning where each input sequence is associated with an extra query, explicitly modeling such conditional contextual dependencies can lead to a more accurate solution, which however cannot be captured by existing self-attention mechanisms. 
In this paper, we propose \textit{conditional self-attention} (CSA), a neural network module designed for conditional dependency modeling.
CSA works by adjusting the pairwise attention between input tokens in a self-attention module with the matching score of the inputs to the given query. Thereby, the contextual dependencies modeled by CSA will be highly relevant to the query. 
We further studied variants of CSA defined by different types of attention.
Experiments on Debatepedia and HotpotQA benchmark datasets show CSA consistently outperforms vanilla Transformer and previous models for the Qsumm problem.%\looseness-1
\end{abstract}

\section{Introduction}

% 1) Contextual dependency is important to a variety of applications.
% 2) Attention has been successfully used to model contextual dependency.
% 3) There are massive applications that need to model contextual dependency conditioned on some query.
% 4) Therefore, we propose conditional self-attention.
% 5) main idea and structure of CSA, experiment result.

Contextual dependency is believed to provide critical information in a variety of NLP tasks. Among the popular neural network structures, convolution is powerful in capturing local dependencies, while LSTM is good at modeling distant relations. Self-attention has recently achieved great success in NLP tasks due to its flexibility of relating two elements in a distance-agnostic manner. For each element, it computes a categorical distribution that reflects the dependency of that element to each of the other elements from the same sequence. The element's context-aware embedding is then computed by weighted averaging of other elements with the probabilities. 
Hence, self-attention is powerful for encoding pairwise relationship into contextual representations.\looseness-1
%Hence, it is effective in re-organizing the sequence by integrating related tokens to produce contextual representations.

However, higher-level language understanding often relies on more complicated dependencies than the pairwise one. One example is the conditional dependency that measures how two elements are related given a premise. In NLP tasks such as query-based summarization and knowledge graph reasoning where inputs are equipped with extra queries or entities, knowing the dependencies conditioned on the given query or entity is extremely helpful for extracting meaningful representations. Moreover, conditional dependencies can be used to build a large relational graph covering all elements, and to represent higher-order relations for multiple ($>$2) elements. Hence, it is more expressive than the pairwise dependency modeled by self-attention mechanisms.

In this paper, we develop \textit{conditional self-attention} as a versatile module to capture the conditional dependencies within an input sequence. CSA is a composite function that applies a cross-attention mechanism inside an ordinary self-attention module. Given two tokens from the input sequence, it first computes a dependency score for each token with respect to the condition by cross-attention, then scales tokens by the computed condition-dependency scores, and applies self-attention to the scaled tokens. In this way CSA is capable of selecting tokens that are highly correlated to the condition, and guiding contextual embedding generation towards those tokens. In addition, when applied to a highly-related token and a loosely-related token, CSA reduces to measuring the global dependency (i.e., single-token importance) of the former. Hence, CSA can capture both local (e.g., pairwise) and global dependency in the same module, and automatically switch between them according to the condition. In contrast, previous works usually need two modules to capture these two types of information.\looseness-1

We then apply CSA to Query-based summarization (Qsumm) task. Qsumm is a challenging task in which accurate retrieval of relevant contexts and thorough comprehension are both critical for producing high-quality summaries. Given a query, it tries to generate from passage a summary pertaining to the query, in either an extractive (i.e., selected from passage) or abstractive (i.e., generated by machine) manner\footnote{Qsumm is related to Question-Answering (QA) but its output contains more background information and explanation than a succinct answer returned by QA.}. We build CSA-Transformer, an attention-only Qsumm model in Section~\ref{sec:csa-transformer} by introducing CSA to Transformer architecture~\citep{vaswani2017attention}. On Debatepedia and HotpotQA\footnote{HotpotQA is originally built for multi-hop QA, but the provided supporting facts can be used as targets of Qsumm.}, CSA-Transformer significantly outperforms baselines for both extractive and abstractive tasks.\looseness-1

\section{Background}
% \vspace{-0.5em}

% \subsection{Word Embedding}

% Word embedding is the basic processing unit in most DNN for sequence modeling. It transfers each discrete token into a representation vector of real values. Given a sequence of tokens (e.g., words or characters) $\bm{w} = [w_1, w_2, \dots, w_n] \in \mathbb{R}^{N \times n}$, where $w_i$ is a one-hot vector, $N$ is the vocabulary size and $n$ is the sequence length. A pre-trained token embedding (e.g. word2vec \citep{mikolov2013distributed}) is applied to $\bm{w}$, which outputs a sequence of low dimensional vectors $\bm{x} = [x_1, x_2, \dots, x_n] \in  \mathbb R ^{d_e \times n}$. This process can be formally written as $\bm{x} = W^{(e)} \bm{w}$, where $W^{(e)} \in \mathbb{R}^{d_e\times N}$ is the embedding weight matrix that can be fine-tuned during the training phase.

\textbf{Attention} Given a sequence $\bm{x} = [x_1, \dots, x_n]$ (e.g., word embeddings) with $x_i\in\mathbb R^{d_e}$ and context vector $c \in \mathbb{R}^{d_c}$ (e.g., representation of another sequence or token), attention \citep{bahdanau2015neural} computes an alignment score between $c$ and each $x_i$ by a compatibility function $f(x_i, c)$. A $\softmax$ function then transforms the alignment scores $\bm{a} \in \mathbb R ^ {n}$ to a categorical distribution $p(z|\bm{x}, c)$ defined as
\begin{align}
\bm{a} &= \left[f(x_i, c)\right]_{i=1}^n \label{eq:tra_attn_1},\\
p(z=i|\bm{x}, c) &= \softmax(\bm{a})[i],~\forall i\in[n] \label{eq:tra_attn_2},
\end{align}
where larger $p(z=i|\bm{x}, c)$ implies that $x_i$ is more relevant to $c$.
A context-aware representation $u$ of $\bm{x}$ is achieved by computing the expectation of sampling a token according to $p(z|\bm{x}, c)$, i.e.,
\begin{equation}\label{equ:tra_attn_output}
u = \sum_{i=1}^n p(z=i|\bm{x},c)x_i=\mathbb E_{i\sim p(z|\bm{x},c)}(x_i).
\end{equation}
Multiplicative attention %(dot-product attention) 
\citep{vaswani2017attention,sukhbaatar2015end,rush2015neural}  and additive attention (multi-layer perceptron attention) \citep{bahdanau2015neural,shang2015neural} are two commonly used attention mechanisms with different compatibility functions $f(\cdot, \cdot)$ defined below respectively,
%\small
\begin{align}
\hspace*{-0.1em} f(x_i, c)&=\left\langle W^{(1)}x_i, W^{(2)}c\right\rangle \label{equ:dot_attn},\\
\hspace*{-0.1em} f(x_i, c)&=w^T\sigma(W^{(1)}x_i+W^{(2)}c + \bm{b}) + b \label{equ:add_attn},
\end{align}
%\normalsize
where $W^{(1)} \!\in\! \mathbb{R}^{d_W \times d_e}, W^{(2)} \!\in\! \mathbb{R}^{d_W \times d_c}$, $w\in\mathbb R^{d_W}$, $\bm{b}$ and $b$ are learnable parameters, and $\sigma(\cdot)$ is an activation function. 
%Additive attention usually achieves better empirical performance than multiplicative attention, but is expensive in time cost and memory consumption. 
% \textbf{Multi-dimensional Attention}: Unlike vanilla attention, in multi-dimensional (multi-dim) attention \citep{shen2017disan}, the alignment score is computed for each feature, i.e., the score of a token pair is a vector rather than a scalar, so the score might be large for some features but small for others. Therefore, it is more expressive than vanilla attention, especially for the words whose meaning varies in different contexts.

\textbf{Self-attention (SA)} is a variant of attention modeling the pairwise relationship between tokens from the same sequence.
% Multi-dim attention has $d_e$ indicators $z_1, \dots, z_{d_e}$ for $d_e$ features. Each indicator has a probability distribution that is generated by applying $\softmax$ to the $n$ alignment scores of the corresponding feature. Hence, for each feature $k$ in each token $i$, we have $P_{ki} \triangleq p(z_k=i|\bm{x}, q)$ where $P \in \mathbb{R}^{d_e \times n}$. A large $P_{ki}$ means that the feature $k$ in token $i$ is important to $q$. The output of multi-dim attention is written as
% \begin{equation}\label{equ:mul_attn_output}
% s = \left[\sum\nolimits_{i=1}^n P_{ki}\bm{x}_{ki}\right]_{k=1}^{d_e}=\left[\mathbb E_{i\sim p(z_k|\bm{x},q)} (\bm{x}_{ki})\right]_{k=1}^{d_e}.
% \end{equation}
% For simplicity, we ignore the subscript $k$ where no confusion is caused. Then, Eq.(\ref{equ:mul_attn_output}) can be rewritten as an element-wise product, i.e., $s = \sum_{i=1}^n P_{\cdot i} \odot x_i$. Here, $P_{\cdot i}$ is computed by the additive attention in Eq.(\ref{eq:add_attn}) where $w^T$ is replaced with a weight matrix $W\in \mathbb{R}^{d_h\times d_e}$, which leads to a score vector for each token pair. 
One line of work a.k.a \textbf{Token2Token self-attention (T2T)} \citep{hu2017mnemonic, shen2017disan} produces context-aware representation for each token $x_j$ based on its dependency to other tokens $x_i$  within $\bm{x}$ by replacing $c$ in Eq.~(\ref{eq:tra_attn_1})-(\ref{equ:tra_attn_output}) with $x_j$. %Thereby, it computes $u_j$ for each token $x_j$ by Eq.~(\ref{eq:tra_attn_1})-(\ref{equ:tra_attn_output}), and the final output $\bm{u} = [u_1, u_2, \dots, u_n]$. 
Notably in Transformer \citep{vaswani2017attention}, a multi-head attention computes $u$ in multiple subspaces of $\bm{x}$ and $c$, and concatenates the results as the final representation.
%Each token $x_j$ is associated with a probability matrix $P^j$ such that $P^j_{ki} \triangleq p(z_k = i | \bm{x}, x_j)$. The output representation for $x_j$ is $s_j = \sum_{i=1}^n P^j_{\cdot i} \odot x_i$ and the final output of token2token self-attention is $\bm{s} = [s_1, s_2, \dots, s_n]$.
\textbf{Source2Token self-attention (S2T)} \citep{lin2017structured,shen2017disan,liu2016learning}, on the other hand, explores the per-token importance with respect to a specific task by removing $c$ from Eq.~(\ref{eq:tra_attn_1})-(\ref{equ:tra_attn_output}). Its output $u$ is an average of all  tokens weighted by their corresponding importance.
%and using compatibility function
% \begin{equation}\label{eq:s2t_attn}
% f(x_i)=w^T\sigma(W^{(1)}x_i + b^{(1)}) + b.
% \end{equation}
%The probability matrix $P$ is defined as $P_{ki} \triangleq p(z_k = i | \bm{x})$. The final output of source2token self-attention has the form $s = \sum_{i=1}^n P_{\cdot i} \odot x_i$
%%%%%%%%%%%%%%%%%%%%%%%%%%%%%%%%%%%%%%%%%%%%%%%%%%%%%%%%%
% \vspace{-5pt}
\section{Conditional Self-Attention}
% \vspace{-5pt}
\begin{figure}[t]
  \vspace{-0pt}
  \begin{center}
  \includegraphics[width=0.45\textwidth]{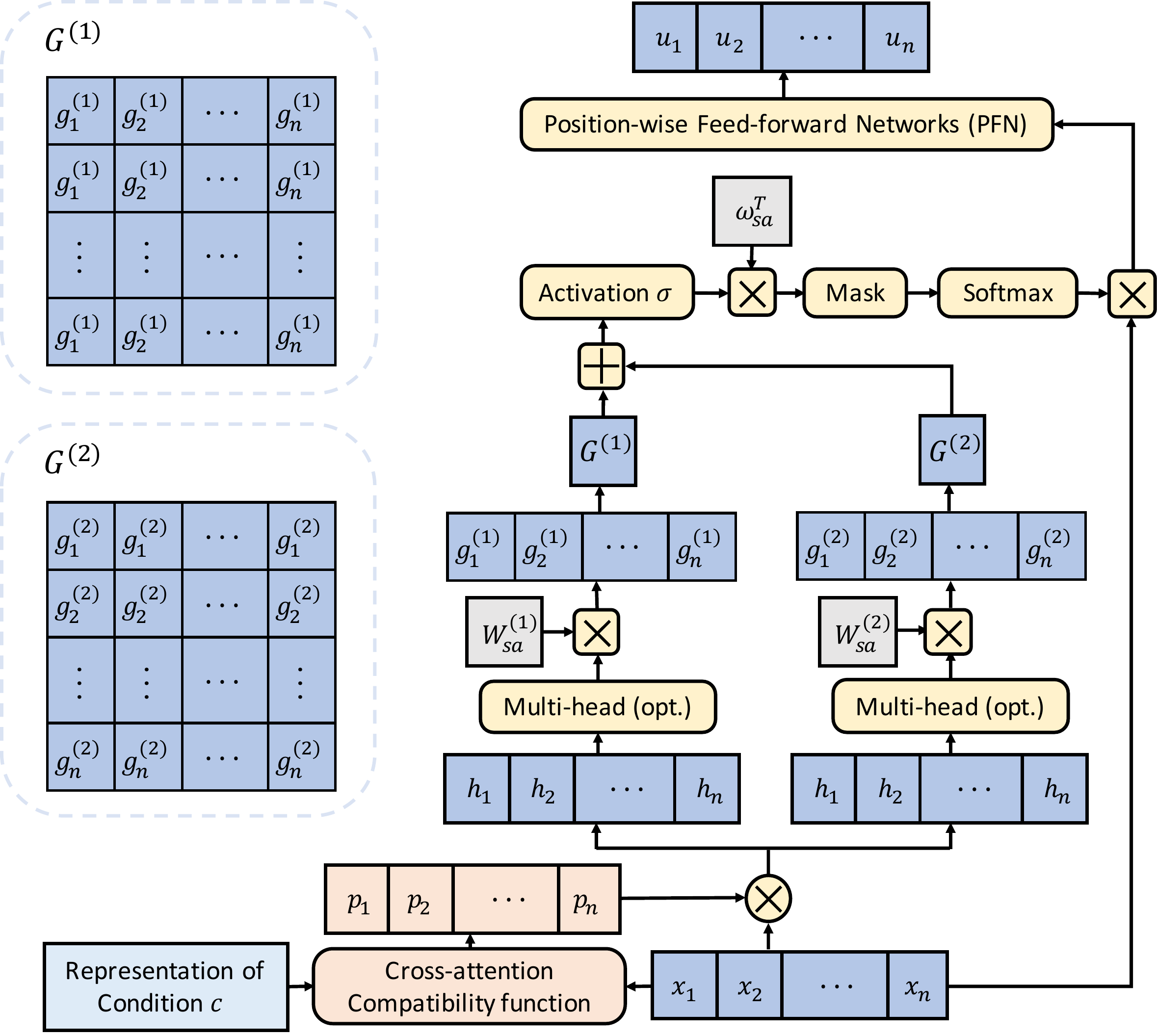} 
  \end{center}
  \vspace{-10pt}
  \caption{CSA module: $\otimes$ is entry-wise multiplication, $\times$ is matrix multiplication, $+$ is entry-wise addition,  $g^{(1)}_i=W^{(1)}_{\text{sa}}h_i$ and $g^{(2)}_j=W^{(2)}_{\text{sa}}h_j$ for $i,j\in[n]$ as in Eq~\ref{equ:csa_addi}. A diagonal mask disables the attention of each token/block to itself before applying $\softmax$.}
  %Here, ``mask'' represents we mask the diagonal entries of the matrix, to prevent that the diagonal entries of the self-attention is much larger than off-diagonal entries
\label{fig:csa}
 \vspace{-10pt}
\end{figure}

Conditional self-attention computes the self-attention score between any two tokens $x_i$ and $x_j$ based on the representation $c$ of a given condition. 

Specifically, to incorporate the conditional information, CSA first applies a cross-attention module (multiplicative or additive attention) to compute the dependency scores of $x_i$ and $x_j$ to $c$ by Eq.~(\ref{eq:tra_attn_1})-(\ref{eq:tra_attn_2}), i.e., $p_i\triangleq p(z=i|\bm{x}, c)$ and $p_j\triangleq p(z=j|\bm{x}, c)$.
Inputs $x_i$ and $x_j$ are then scaled by $p_i$ and $p_j$ to obtain $h_i\triangleq p_i x_i$ and $h_j\triangleq p_j x_j$. Finally, an additive self-attention is applied to $h_i$ and $h_j$ with compatibility function 
\begin{equation}\label{equ:csa_addi}
\begin{array}{ll}
\hspace*{-0.1em}f_{\text{csa}}(x_i, x_j| c)\triangleq\\
\hspace*{-0.1em}~~~~~~w^T_{\text{sa}}\sigma(W^{(1)}_{\text{sa}}h_i+W^{(2)}_{\text{sa}}h_j + \bm{b}_{\text{sa}}) + b_{\text{sa}},
\end{array}
\end{equation}
%Then, the context-aware embedding $u_j$ for $h_j$ is computed in similar manner as Eq.~(\ref{eq:tra_attn_1})-(\ref{equ:tra_attn_output}), i.e.,
resulting in context-aware embedding $u_j$ for $x_j$:
\begin{equation}\label{equ:csa_agg}
    u_j = \sum_{i=1}^n \softmax(\left[f_{\text{csa}}(h_i, h_j)\right]_{i=1}^n)[i]\times h_i.
\end{equation}

We extend the above model by using multi-head mechanism and position-wise feed-forward networks (PFN) proposed in~\citet{vaswani2017attention}: Eq.~(\ref{equ:csa_addi})-(\ref{equ:csa_agg}) are applied to $K$ subspaces of $\bm{h}$ from $[\Theta^{(k)}\bm{h}]_{k=1}^K$, whose outputs $[u^{(k)}_j]_{k=1}^K$ are then concatenated and processed by a linear projection (with $w^{\text{head}}\in\mathbb R^K$) and a PFN layer
\begin{equation}
\hspace*{-0.2em} u_j = \text{PFN}\left(\left[u^{(1)}_j; u^{(2)}_j; \cdots; u^{(K)}_j\right] w^{\text{head}}\right).
\end{equation}
\section{Query-based Summarization Model}\label{sec:csa-transformer}
% \vspace{-5pt}
% In Qsumm, each data instance is a triple $(p,q,s)$, where $p$ is the passage, $q$ is the query, and $s$ is the summarization of the passage conditioned on the query, that is, $s$ largely depends on the contextual information given in $q$, because a) it provides information that which part of the passage $p$ the summarization $s$ should emphasis on, and b) it does not directly contribute to the summarization, since the summarization $s$ still covers the key points of the passage, not just answering the query.

In {\it Qsumm}, each data instance is a triplet $(\bm{x},\bm{q},\bm{s})$, representing passage, query, and a summary of passage conditioned on query. Summary $\bm{s}$ should not only focus on the content in $\bm{x}$ that is most relevant to $\bm{q}$, but it also needs to be a complete summary rather than a short answer to $\bm{q}$ as opposed to QA. In other words, it should cover any background information and related key points about $\bm{q}$ if they are also covered by $\bm{x}$.\looseness-1

\begin{figure}[t]
%\vspace{-1pt}
\begin{center}
%\framebox[4.0in]{$\;$}
\includegraphics[width=0.43\textwidth]{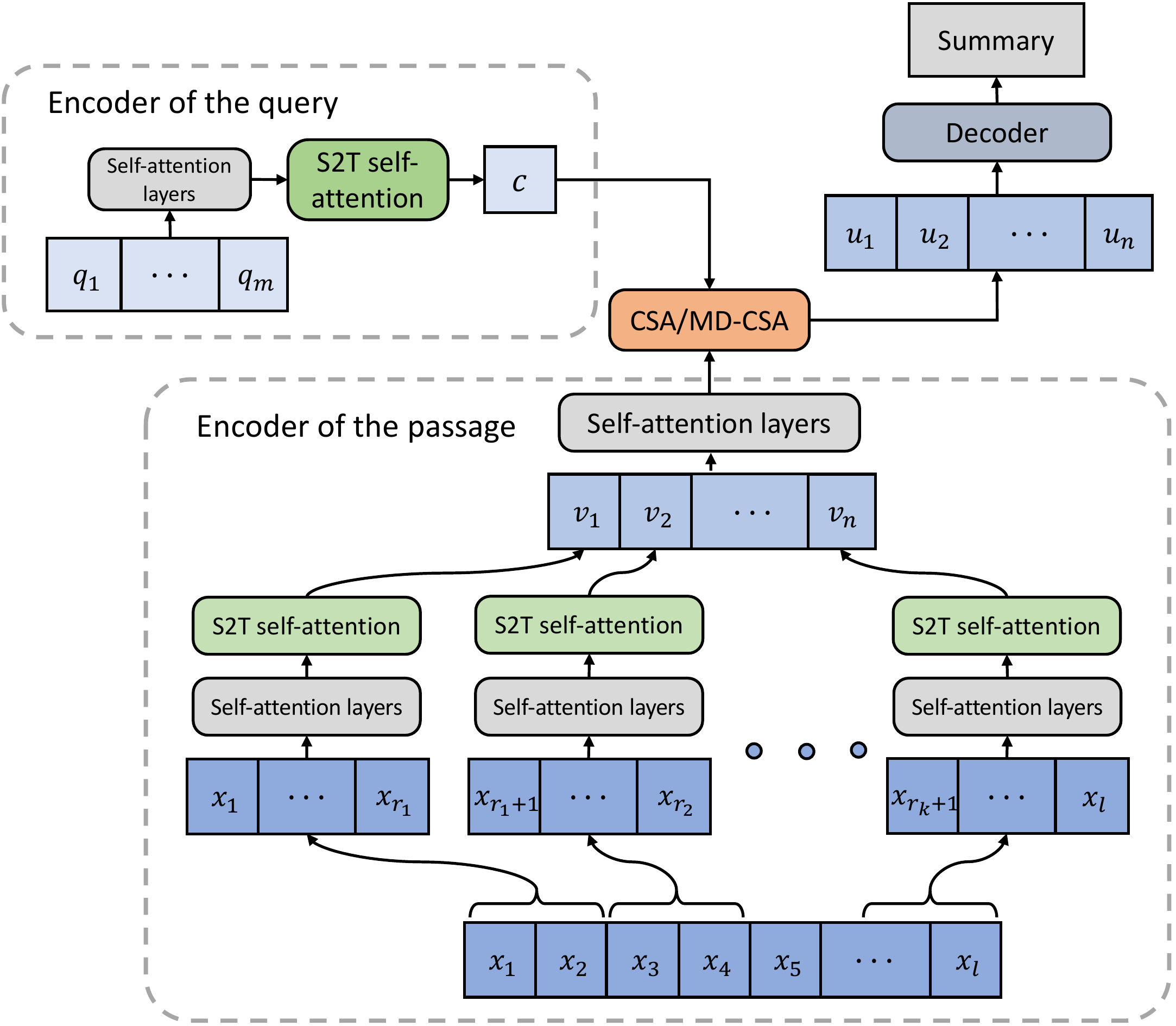} 
\end{center}
\vspace{-10pt}
\caption{CSA-Transformer with block self-attention.}
\label{fig:csan}
\vspace{-10pt}
\end{figure}

%We develop a {\it Transformer}-style neural network built from {\it CSA} and self-attention (SA) modules for {\it Qsumm}. Its architecture is illustrated in Figure~\ref{fig:csan}, which consists of the following four parts.
Figure~\ref{fig:csan} illustrates our Transformer-style neural network for Qsumm. It is built with CSA modules containing two encoders, a CSA layer and a decoder, which are elaborated below.
%In this section we use query-based summarization as an example, to introduce how the proposed CSA module is used in the network to provide contextual information. 

%Figure \ref{fig:csan} shows the basic architecture of the proposed network. The network consists of the following four modules. 

\textbf{Encoder of passage $\bm{x}$} 
%The passage in a summarization task can be very long. For example, in the HotpotQA dataset we use, the passages are on average more than 800 tokens. Therefore, for long passages, 
We adopt hierarchical block self-attention~\citep{shen2018bi, wikisumm} for the sake of memory efficiency when processing long passages by splitting $\bm{x}$ into $n$ blocks (subsequences). For each block, we apply a few Transformer encoding layers with shared parameters across the blocks, followed by a S2T self-attention layer that compresses the subsequence into a compact vector. Thereby, we get a shorter sequence of $n$ vectors, on which we apply a few more Transformer encoding layers to obtain the final sequence $\bm{v}$.

\textbf{Encoder of query $\bm{q}$} Given a query $\bm{q}$ as a sequence of tokens, we apply a few Transformer encoding layers, followed by an S2T self-attention layer, to produce a vector $c$ as representation of the condition in CSA.

\textbf{Conditional self-attention layer} Given the encoded query $c$ and the encoded passage ${\bm v}$, a conditional self-attention module as in Figure~\ref{fig:csa} (with input ${\bm x}={\bm v}$) produces a sequence of context-and-condition-aware representations ${\bm u}$. %Then, we apply a source2token self-attention layer to $u$ and achieve a compact representation vector $u'$.

\textbf{Decoder} Depending on whether the output summary is abstractive or extractive, we apply different decoders to ${\bm u}$. We will elaborate which decoder is used for each dataset in the experiments.

% Depending on different output form, different kinds of decoder should be adopted. For abstractive summarization, we need to generate new sentences for summarization instead of using the existing sentences in the passage. Here, we adopt a transformer decoder, together with cross-entropy loss function. 

% For extractive summarization, where we just need to extract sentences from the passage, the decoder can be much simpler. The output of the CSA module is a sequence of vectors. For each vector, we first project it into a scalar, and the 

%%%%%%%%%%%%%%%%%%%%%%%%%%%%%%%%%%%%%%%%%%%%%%%%%%%%%%%%%%%%%%%%%%%%%%%
\begin{table*}[h]
\caption {Query-based summarization on Debatepedia (abstractive) and HotpotQA (extractive). Two CSA models are evaluated: (Mul) and (Add) refer to multiplicative and additive cross-attention used in CSA.} \label{tab:result}
\vspace{-8pt}
\fontsize{9.8}{12}\selectfont
\centering
\begin{tabular}{lcccccc}
\noalign{\hrule height 1pt}
\multirow{2}{*}{Model} & \multicolumn{3}{c}{Debatepedia~\citep{nema2017diversity}}& \multicolumn{3}{c}{HotpotQA~\citep{yang2018hotpotqa}}\\
\cline{2-7}
          & {\sc Rouge}-1 & {\sc Rouge}-2 & {\sc Rouge}-L  & {\sc Rouge}-1 & {\sc Rouge}-2 & {\sc Rouge}-L       \\ 
\noalign{\hrule height 1pt}
Transformer~\cite{vaswani2017attention}     & 28.16   &  17.48       &   27.28  & 35.45   &  28.17     & 30.31 \\ 
%\cline{1-7}
UT~\cite{dehghani2018universal} &36.21    &  26.75  &    35.53       & 41.58  &  32.28    & 34.88 \\ 
SD2~\cite{nema2017diversity}  &   41.26    & 18.75   &  40.43 & -- & -- & --  \\ %
%\cline{1-7}
CONCAT    & 41.72 & 33.62 & 41.25 & 28.23 & 24.50 & 24.76  \\ %
%\cline{1-7}
ADD & 41.10 & 33.35 & 40.72 & 32.84 & 28.01 & 28.53  \\ %
%\cline{1-7}
% Simple-MLP  &  -- & -- & -- & 17.16 & 14.82 & 15.08  \\ %
%\cline{1-7}
\noalign{\hrule height 1pt}
% Query-CSA,Multi.,Score & \textbf{45.20}   & \textbf{35.98}   & \textbf{44.69}   \\ 
% \cline{1-4}
CSA Transformer (Mul) & 41.70  & 32.92   & 41.29  & \textbf{59.57}   & \textbf{49.89}   & \textbf{48.34} \\ 
%\cline{1-7}
% MD-CSA Transformer (Mul)& 42.36   & 32.84   & 41.69 & 47.22   & 38.61   & 38.10  \\ 
%\cline{1-7}
CSA Transformer (Add) & \textbf{46.44}   & \textbf{37.38}   & \textbf{45.85} & 47.00   & 37.78   & 39.52   \\ 
%\cline{1-7}
% MD-CSA Transformer (Add) & 43.00   & 34.11   & 42.52 &48.12   & 38.59  & 40.14  \\ 
\noalign{\hrule height 1pt}
\end{tabular}
\vspace{-10pt}
\end{table*}
%%%%%%%%%%%%%%%%%%%%%%%%%%%%%%%%%%%%%%%%%%%%%%%%%%%%%%%%%%%%%%%%%%%%%%%

%\vspace{-5pt}
\section{Experiments}
% \vspace{-5pt}
We evaluate CSA-Transformer for abstractive summarization task on Debatepedia~\citep{nema2017diversity} and extractive summarization task on HotpotQA~\citep{yang2018hotpotqa}, with different decoders applied. We consider two variants that use CSA module with additive or multiplicative cross-attention. 
%In the experiments, we consider the query based summarization model with four kinds of CSA layer: CSA, Diag-CSA with additive and multiplicative cross attention, respectively. 
We first compare them with baselines that does not consider queries, namely \textit{Transformer}~\cite{vaswani2017attention} and Universal Transformer (\textit{UT})~\cite{dehghani2018universal}.
We further compare them with the following baselines: 

\textit{SD2}: See \citet{nema2017diversity}; 

\textit{CONCAT}: Concatenate the query and the passage, and feed it into Transformer; 

\textit{ADD}: Add the query encoded vector to every word’s word embedding from the passage encoder, and feed it to the decoder, i.e., $u_i = v_i+c$; 

% \textit{Simple-MLP}: Compute CSA score by a simple CSA layer, i.e.,
% \begin{align*}
%     &f_{\text{csa}}(x_i,x_j|c)=  \\
%     & ~~~w_{\text{mlp}}*\sigma(W^{(1)}_{\text{mlp}}*x_i+W^{(2)}_{\text{mlp}}*x_j+W^{(3)}_{\text{mlp}}*c).
% \end{align*}

Note that Debatepedia dataset is also tested in \citet{baumel2018query}. However, as claimed in Section 5.4 of \citet{baumel2018query}, their model targets a different setting and yields summaries ten times longer than required on this dataset. Therefore, their result is not directly comparable with ours.

% : UT is a generalization of Transformer. It is a parallel-in-time self-attentive recurrent sequence model.

\subsection{Abstractive Query-based Summarization}

\noindent\textbf{Debatepedia}~\citep{nema2017diversity} is crawled from 663 debates of 53 categories such as Politics, Law, Crime, etc. It has 12000 training data, 719 validation data, and 1000 test data. Examples of the data instances can be found in Appendix.

\begin{figure}[t]
% \vspace{-10pt}
\begin{center}
%\framebox[4.0in]{$\;$}
\includegraphics[width=0.48\textwidth]{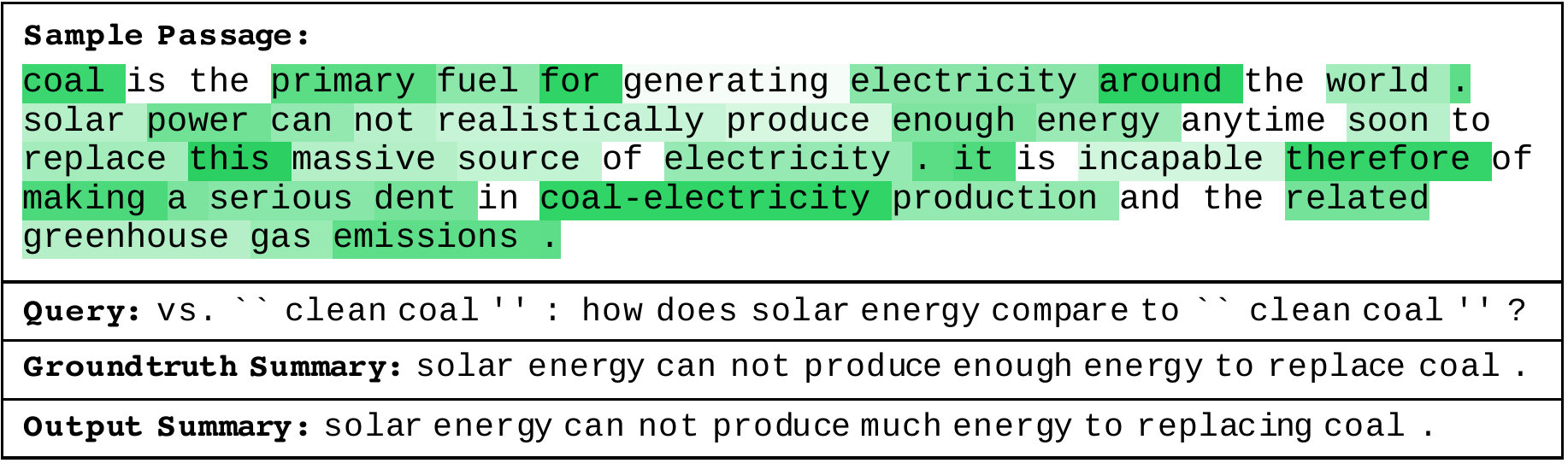}
\end{center}
\vspace{-10pt}
\caption{An example of query-based summarization on Debatepedia: the passage is highlighted according to the cross-attention scores $p_i$, and the abstractive summary is produced by CSA Transformer (Add).}
\label{fig:sample}
\vspace{-15pt}
\end{figure}

\noindent\textbf{Setup} Since the passages in Debatepedia are relatively short (66.4 tokens per instance on average), we use a Transformer encoding layer as the passage encoder without splitting the passage into blocks. Transformer decoding layer is used for decoding. Please refer to Appendix for more details.

\noindent\textbf{Result}
Table~\ref{tab:result} shows both two CSA models consistently achieve better {\sc Rouge} scores~\citep{lin2004automatic} than the baselines. 
% MD-CSA does not outperform CSA since it uses the same hyper-parameters tuned for CSA.  
Note our models have much higher {\sc Rouge}-2 scores than baselines, which suggests the summarization generated by CSA is more coherent.
Figure~\ref{fig:sample} is an example of CSA summary, with the passage highlighted by cross-attention score $p_i$. The learned attention scores emphasize not only lexical units such as "coal-electricity" but also conjunctive adverb such as "therefore." More example summaries can be found in Appendix.

% \begin{table}[]
% \caption {Hotpot (Validation Set)} \label{tab:hotpot} 
% \begin{tabular}{l|llll}
% \cline{1-5}
% Model           & Exact Match & F$_1$ & Recall@5 & Recall@10        \\ 
% \cline{1-5}
% \citep{yang2018hotpotqa}     &   5.28     & 41.0   &  --       & -- \\ 
% \cline{1-5}
% QFE    &   14.20     & 44.4   &     --    & -- \\ 
% \cline{1-5}
% Transformer      &  7.30 (3.45)      & 76.3 (21.0)   &  36.5    &   54.7 \\ 
% \cline{1-5}
% Query-CSA model & 15.0 (9.76)     & 80.2 (49.3)    &  53.4   & 71.4 \\ 
% \cline{1-5}
% \end{tabular}
% \end{table}

% \begin{table}[]
% \caption {CNN/Daily Mail} \label{tab:cnn} 
% \begin{tabular}{l|llll}
% \cline{1-4}
% Model           & Bleu-1 & Rouge-1 & Rouge-L        \\ 
% \cline{1-4}
% Transformer     &         &         & 18.9/17.4/17.7 \\ 
% \cline{1-4}
% Query-CSA model & 28.6        &         & 26.7/22.0/23.8/25.6 \\ 
% \cline{1-4}
% \end{tabular}
% \end{table}

% \begin{figure}[t]
% % \vspace{-10pt}
% \begin{center}
% %\framebox[4.0in]{$\;$}
% \includegraphics[width=0.45\textwidth]{sample_attn.png} 
% \end{center}
% \vspace{-10pt}
% \caption{\label{fig:sample_attn} Sample passages with attention. }
% \end{figure}
\vspace{-1pt}
\subsection{Extractive Query-based Summarization}
\vspace{-1pt}
\noindent\textbf{HotpotQA}~\citep{yang2018hotpotqa} is a multi-hop QA dataset including 113k Wikipedia-based instances, each of which consist of four parts: a question (query), a context, a short answer, and support facts that the answer is extracted from. The support facts are sentences selected from the context and contain reasonably more information of the context than the answer. 

We adapt HotpotQA as a benchmark for extractive query-based summarization by removing the answer of each instance and treating the context as passage ${\bm x}$, the question as query $\bm q$, and the support facts as summary ${\bm s}$ in {\it Qsumm}. Since the test set has not been released, in order to compute {\sc Rouge} score in evaluation, we randomly split the original training set to a training set of 85564 instances and a validation set of 5000 instances, and use the original validation set as our test set. 

\noindent\textbf{Setup} We split the passage into blocks, each to be one of its sentences. 
%We adopt 2 transformer encoding layers for each block, and 2 transformer encoding layers for the concatenated blocks after the source2token layer. 
%We also adopt 2 transformer encoding layer for the transformer encoding module. 
The decoder is 2 Transformer decoder layers, whose output is a sequence of vectors,  followed by a projection to a sequence of scalars, each representing a sentence. Each scalar is processed by a $\sigmoid$ function for binary classification deciding whether the sentence should be selected. We use cross-entropy as our loss function. Please refer to Appendix for more details.

\noindent\textbf{Result} Both CSA models achieve significantly better {\sc Rouge} scores than the baselines in Table~\ref{tab:result}. Different from Debatepedia results, CSA module with multiplicative cross-attention performs better than the additive one. This is because in each dataset we adopt the same hyper-parameters for both variants -- in Debatepedia the hyper-parameters favor multiplicative cross-attention, while in HotpotQA they prefer the additive one. The experimental results show that both variants can have significantly better performance than the baselines. It also suggests that we cannot simply adopt the hyper-parameters from another tasks without fine-tuning -- although the model can be more compelling than the baselines if we do so.

% \begin{table*}[]
% \caption {Hotpot (not retrained yet)} \label{tab:hotpot} 
% \begin{tabular}{l|llll}
% \cline{1-4}
% Model           & Rouge-1 & Rouge-2 & Rouge-L        \\ 
% \cline{1-4}
% Transformer     & 35.45   &  28.17     & 30.31 \\ 
% \cline{1-4}
% % Transformer+BiDAF    & 59.08   &     49.43    & 48.88 \\ 
% % \cline{1-4}
% Query-CSA,Multi.,Score & \textbf{59.57}   & \textbf{49.89}   & \textbf{48.34}   \\ 
% \cline{1-4}
% Query-CSA,Multi.,Diag & \textbf{}   & \textbf{}   & \textbf{}   \\ 
% \cline{1-4}
% Query-CSA,Addi.,Score & 47.00   & 37.78   & 39.52   \\ 
% \cline{1-4}
% Query-CSA,Addi.,Diag (running) &43.13   & 34.76  & 36.06   \\ 
% \cline{1-4}
% \end{tabular}
% \end{table*}

% \begin{table}[]
% \caption {Hotpot (Validation Set)} \label{tab:hotpot} 
% \begin{tabular}{l|ll}
% \cline{1-3}
% Model           & Exact Match & F$_1$        \\ 
% \cline{1-3}
% \citep{yang2018hotpotqa}     &  20.32 & 64.49   \\ 
% \cline{1-3}
% QFE    &   57.75 & 84.49  \\ 
% \cline{1-3}
% Transformer (val) &  11.02  &    40.30     \\ 
% \cline{1-3}
% Query-CSA model (val) & 24.62 & 61.82 \\ 
% \cline{1-3}
% \end{tabular}
% \end{table}
%\vspace{-5pt}
\vspace{-2pt}
\section{Conclusion}
\vspace{-5pt}
This paper introduces conditional self-attention (CSA) and its variants as versatile and plugin modules for conditional contextual dependency modeling. 
We develop an attention-only neural network built from CSA and Transformer for query-based summarization. 
It consistently outperforms vanilla Transformer and other baselines for abstractive and extractive Qsumm tasks on Debatepedia and HotpotQA datasets.

Due to its capability of modeling conditional dependencies, CSA can be naturally applied to tasks defined on graph structure, e.g., classification and clustering on graph(s), logical reasoning, knowledge graph reasoning, to name a few. To handle multiple queries, we can apply a CSA module per query and combining their outputs by an S2T self-attention module.

\bibliography{emnlp-ijcnlp-2019}
\bibliographystyle{acl_natbib}

% \hl{To Yujia:}
% \begin{enumerate}
    % \item \hl{Change all the double quote symbols "***" to ``***''; } {\color{blue} Changed. Left the ones copied from data.}
    % \item \hl{Change all the png figures to pdf formats (from Figure 3 to Figure 7 only, Figure 1 and Figure 2 have been transferred to pdf) by exporting pdf from ppt$\rightarrow$cropping the target figure on its page$\rightarrow$deleting other pages and save. In case the resulted pdf has an unacceptable large size, we can use png;} {\color{red} Not working. A weird box would appear around the image for passage inside the table. See sample.pdf for example. Any ideas how to remove it?}{\color{blue} I think it is caused by the non-vector graphs of the original pictures. Maybe trying to directly output pdf figure from matplotlib or other tools.}
    % \item \hl{Change all the outdated arXiv papers in reference to its accepted version if accepted by somewhere (I changed some, but suspected that there are still some outdated);}
    % \item \hl{``masked-softmax'' in Figure 1 does not have a definition: we can either place the definition in Appendix or explain it in main paper if simple to explain. I guess it's just a column-wise softmax, is it?} {\color{blue} It's just mask+softmax. What about now?}
% \end{enumerate}

\onecolumn
\appendix

% \section{Appendices}
% \label{sec:appendix}
\vspace{10pt}
\centerline{\Large{\textbf{Appendix}}}
\vspace{15pt}

\section{Related Works}
\textbf{Attention} Attention mechanisms \citep{bahdanau2015neural, sutskever2014sequence, P18-1224}, especially self-attention mechanisms have achieved great success on variety of NLP tasks \cite{vaswani2017attention,devlin2018bert,C18-1154,radford2019language}. Our work can be viewed as an extension of Transformer \cite{vaswani2017attention}. Also, our work adapts the source to token layers (S2T) from \citet{lin2017structured} and \citet{shen2017disan}.

We notice that the BERT~\citep{devlin2018bert} model (released after we achieved our main results) can implicitly capture conditional dependency by taking the concatenation of the passage and query with a separation symbol as input. However, CSA is specifically designed for modeling conditional dependency explicitly, and can potentially improve Transformer, BERT, GPT-2~\cite{radford2019language} or other models trained for tasks such as Qsumm.

\noindent\textbf{Query-based Summarization} Many works are devoted to do summarization with attention mechanism \citep{sutskever2014sequence,tan2017abstractive}, but few considers the conditional dependency on query. Many works are devoted to do query-based summarization \cite{wan2008multi, zhao2009using, park2006query, schilder2008fastsum}, but few adopts deep learning based methods. \citet{hasselqvist2017query} and \citet{cao2016attsum} propose to do query-based summarization using deep networks with different setting -- in their setting, the query is a key word, instead of a question. A very related work is \citet{nema2017diversity}. Their setting is the same as us, and crawled Debatepedia dataset.

\section{Implementation Detail}

In this section, we elaborate more detail on the experiments.

\subsection{Final Settings of the Experiments}

\begin{table}[H]
\caption {Setting of the tests.} \label{tab:setting} 
\begin{tabular}{lll}
%\cline{1-3}
%\cline{1-3}
\toprule
Dataset           & Debatepedia & HotpotQA      \\ 
\cline{1-3}
Pre-trained embedding     &  None & \texttt{glove.840B.300d.txt}    \\ 
%\cline{1-3}
Batch size    & 64  & 32  \\ 
%\cline{1-3}
Embedding dimension &  128  &    300     \\ 
%\cline{1-3}
Hidden dimension & 128 & 300 \\ 
%\cline{1-3}
CSA inner dimension& 16 & 16 \\ 
%\cline{1-3}
\#heads in CSA layer & 4 & 4 \\ 
%\cline{1-3}
\#SA layers before block & 1 & 2 \\ 
%\cline{1-3}
\#SA layers after block & 0 & 2 \\ 
%\cline{1-3}
\#SA layers in query & 1 & 2 \\ 
%\cline{1-3}
\#SA layers after CSA & 1 & 2 \\ 
%\cline{1-3}
\#heads in SA layer & 4 & 4 \\ 
%\cline{1-3}
Loss & Cross-entropy & Cross-entropy (binary) \\ 
Dropout & 0.2 & 0.1 \\
%\cline{1-3}
Learning rate & 0.5 & $10^{-4}$ \\ 
%\cline{1-3}
Learning rate decay & Expo. with factor 0.8 per 3000 iterations & Expo. with factor 0.9 per epoch \\ 
% \cline{1-3}
% \cline{1-3}
\bottomrule
\end{tabular}
\end{table}

In every experiment, we pad the sequence $\bm{x}$ and query $\bm{q}$ in each batch to have the same length, for the convenience of parallelization. Correspondingly, a mask for the padded entries is used in every layers of the model. 

Also, for HotpotQA, we use binary classification to decide whether this sentence is picked. In order to do that, we need to pick a threshold -- if the classification score of this sentence is larger than the threshold, we pick this sentence; if not, we do not pick it. This threshold is set to be 0.25. 

We also use pointing mechanism and coverage mechanism \citep{see2017get} for abstractive summarization.

\subsection{Tuning Process}

For both dataset, the following parameters are tuned: hidden dimension from 64 to 300, CSA inner dimension from 16 to 64, number of layers in each part from 0 to 3, learning rate. We remark the following two aspects largely affect the performance:

\begin{enumerate}
    \item The model size. The two datasets are different in size. Debatepedia is smaller. To achieve a good perforamce, the embedding dimension, hidden dimension and number of layers are smaller. HotpotQA is larger, so that we set the embedding dimension, hidden dimension and number of layers to be also larger.
    \item The learning rate. We notice that usually smaller learning rate can converge to better result than larger ones, although may be slower. 
\end{enumerate}

For Debatepedia, the performance of CSA Transformer (Add) is better than other three. For HotpotQA, the performance of CSA Transformer (Mul) is better than other three by a large margin. This is because in the tuning process, we fix the kind of CSA layer, and tune the other parameters. After we get a good set of other parameters, we use the setting for three other kinds of CSA layers. That is why one setting of CSA layer is much better than the other three. The other three settings are not extensively tuned than the one with better performance. 

\subsection{Implementation of the baselines}

The results of two baselines, Transformer \citep{vaswani2017attention} and Universal Transformer \citep{dehghani2018universal} is run by us. The implementation of Transformer model and some of the parameter setting is largely borrowed from \texttt{https://github.com/jadore801120/attention-is-all-you-need-pytorch}. For both datasets, we use the same number of layers in each part. The implementation of Universial Transformer model and some of the parameter setting is largely borrowed from \texttt{https://github.com/andreamad8/Universal-Transformer-Pytorch}. For both datasets, since Adaptive Computation Time (ACT) is used, we only use one layer for each ``SA layers'' block in the encoder part. For decoder, we use exactly the same decoder for both Transformer and Universal Transformer as CSA based methods. We also remark that the structure of our implementation is adapted from \texttt{https://github.com/OpenNMT/OpenNMT-py}.
% \vspace{50pt}
\section{Summary Examples}

In this section, we show a few summarization examples.

\subsection{HotpotQA}

\begin{longtable}[H]{|p{15.5cm}|}
% \begin{tabular}{|p{16cm}|}
\hline

Example 1 \\ 
\noalign{\hrule height 1pt}
\textbf{Passage:}     \\   Ed Wood is a 1994 American biographical period comedy-drama film directed and produced by Tim Burton, and starring Johnny Depp as cult filmmaker Ed Wood. The film concerns the period in Wood's life when he made his best-known films as well as his relationship with actor Bela Lugosi, played by Martin Landau. Sarah Jessica Parker, Patricia Arquette, Jeffrey Jones, Lisa Marie, and Bill Murray are among the supporting cast.
\\
Scott Derrickson (born July 16, 1966) is an American director, screenwriter and producer. He lives in Los Angeles, California. He is best known for directing horror films such as ``Sinister'', ``The Exorcism of Emily Rose'', and ``Deliver Us From Evil'', as well as the 2016 Marvel Cinematic Universe installment, ``Doctor Strange.''
\\
Woodson is a census-designated place (CDP) in Pulaski County, Arkansas, in the United States. Its population was 403 at the 2010 census. It is part of the Little Rock–North Little Rock–Conway Metropolitan Statistical Area. Woodson and its accompanying Woodson Lake and Wood Hollow are the namesake for Ed Wood Sr., a prominent plantation owner, trader, and businessman at the turn of the 20th century. Woodson is adjacent to the Wood Plantation, the largest of the plantations own by Ed Wood Sr.
\\
Tyler Bates (born June 5, 1965) is an American musician, music producer, and composer for films, television, and video games. Much of his work is in the action and horror film genres, with films like ``Dawn of the Dead, 300, Sucker Punch,'' and ``John Wick.'' He has collaborated with directors like Zack Snyder, Rob Zombie, Neil Marshall, William Friedkin, Scott Derrickson, and James Gunn. With Gunn, he has scored every one of the director\'s films; including ``Guardians of the Galaxy'', which became one of the highest grossing domestic movies of 2014, and its 2017 sequel. In addition, he is also the lead guitarist of the American rock band Marilyn Manson, and produced its albums ``The Pale Emperor'' and ``Heaven Upside Down''.
\\
Edward Davis Wood Jr. (October 10, 1924 – December 10, 1978) was an American filmmaker, actor, writer, producer, and director.
\\
Deliver Us from Evil is a 2014 American supernatural horror film directed by Scott Derrickson and produced by Jerry Bruckheimer. The film is officially based on a 2001 non-fiction book entitled ``Beware the Night'' by Ralph Sarchie and Lisa Collier Cool, and its marketing campaign highlighted that it was ``inspired by actual accounts''. The film stars Eric Bana, Édgar Ramírez, Sean Harris, Olivia Munn, and Joel McHale in the main roles and was released on July 2, 2014.
\\
Adam Collis is an American filmmaker and actor. He attended the Duke University from 1986 to 1990 and the University of California, Los Angeles from 2007 to 2010. He also studied cinema at the University of Southern California from 1991 to 1997. Collis first work was the assistant director for the Scott Derrickson\'s short ``Love in the Ruins'' (1995). In 1998, he played ``Crankshaft'' in Eric Koyanagi\'s ``Hundred Percent''.
\\
Sinister is a 2012 supernatural horror film directed by Scott Derrickson and written by Derrickson and C. Robert Cargill. It stars Ethan Hawke as fictional true-crime writer Ellison Oswalt who discovers a box of home movies in his attic that puts his family in danger.
\\
Conrad Brooks (born Conrad Biedrzycki on January 3, 1931 in Baltimore, Maryland) is an American actor. He moved to Hollywood, California in 1948 to pursue a career in acting. He got his start in movies appearing in Ed Wood films such as ``Plan 9 from Outer Space'', ``Glen or Glenda'', and ``Jail Bait.'' He took a break from acting during the 1960s and 1970s but due to the ongoing interest in the films of Ed Wood, he reemerged in the 1980s and has become a prolific actor. He also has since gone on to write, produce and direct several films.
\\
Doctor Strange is a 2016 American superhero film based on the Marvel Comics character of the same name, produced by Marvel Studios and distributed by Walt Disney Studios Motion Pictures. It is the fourteenth film of the Marvel Cinematic Universe (MCU). The film was directed by Scott Derrickson, who wrote it with Jon Spaihts and C. Robert Cargill, and stars Benedict Cumberbatch as Stephen Strange, along with Chiwetel Ejiofor, Rachel McAdams, Benedict Wong, Michael Stuhlbarg, Benjamin Bratt, Scott Adkins, Mads Mikkelsen, and Tilda Swinton. In ``Doctor Strange'', surgeon Strange learns the mystic arts after a career-ending car accident.
\\ \hline
\textbf{Query:}     Were Scott Derrickson and Ed Wood of the same nationality?       \\ \hline
\textbf{Ground Truth:}   Scott Derrickson (born July 16, 1966) is an American director, screenwriter and producer. Edward Davis Wood Jr. (October 10, 1924 – December 10, 1978) was an American filmmaker, actor, writer, producer, and director.     \\ \hline
\textbf{CSA Summary:}   Scott Derrickson (born July 16, 1966) is an American director, screenwriter and producer. Edward Davis Wood Jr. (October 10, 1924 – December 10, 1978) was an American filmmaker, actor, writer, producer, and director.     \\
\hline
% \end{tabular}
\end{longtable}

\newpage
\begin{longtable}[H]{|p{15.5cm}|}
% \begin{tabular}{|p{15.5cm}|}
\hline
Example 2 \\ 
\noalign{\hrule height 1pt}
\textbf{Passage:}  \\  Meet Corliss Archer, a program from radio\'s Golden Age, ran from January 7, 1943 to September 30, 1956. Although it was CBS\'s answer to NBC\'s popular ``A Date with Judy'', it was also broadcast by NBC in 1948 as a summer replacement for ``The Bob Hope Show''. From October 3, 1952 to June 26, 1953, it aired on ABC, finally returning to CBS. Despite the program\'s long run, fewer than 24 episodes are known to exist.
\\
Shirley Temple Black (April 23, 1928 – February 10, 2014) was an American actress, singer, dancer, businesswoman, and diplomat who was Hollywood's number one box-office draw as a child actress from 1935 to 1938. As an adult, she was named United States ambassador to Ghana and to Czechoslovakia and also served as Chief of Protocol of the United States.
\\
Janet Marie Waldo (February 4, 1920 – June 12, 2016) was an American radio and voice actress. She is best known in animation for voicing Judy Jetson, Nancy in ``Shazzan'', Penelope Pitstop, and Josie in ``Josie and the Pussycats'', and on radio as the title character in ``Meet Corliss Archer''.
\\
Meet Corliss Archer is an American television sitcom that aired on CBS (July 13, 1951 - August 10, 1951) and in syndication via the Ziv Company from April to December 1954. The program was an adaptation of the radio series of the same name, which was based on a series of short stories by F. Hugh Herbert.
\\
The post of Lord High Treasurer or Lord Treasurer was an English government position and has been a British government position since the Acts of Union of 1707. A holder of the post would be the third-highest-ranked Great Officer of State, below the Lord High Steward and the Lord High Chancellor.
\\
A Kiss for Corliss is a 1949 American comedy film directed by Richard Wallace and written by Howard Dimsdale. It stars Shirley Temple in her final starring role as well as her final film appearance. It is a sequel to the 1945 film ``Kiss and Tell''. ``A Kiss for Corliss'' was retitled ``Almost a Bride'' before release and this title appears in the title sequence. The film was released on November 25, 1949, by United Artists.
\\
Kiss and Tell is a 1945 American comedy film starring then 17-year-old Shirley Temple as Corliss Archer. In the film, two teenage girls cause their respective parents much concern when they start to become interested in boys. The parents' bickering about which girl is the worse influence causes more problems than it solves.
\\
The office of Secretary of State for Constitutional Affairs was a British Government position, created in 2003. Certain functions of the Lord Chancellor which related to the Lord Chancellor's Department were transferred to the Secretary of State. At a later date further functions were also transferred to the Secretary of State for Constitutional Affairs from the First Secretary of State, a position within the government held by the Deputy Prime Minister.
\\
The Village Accountant (variously known as ``Patwari'', ``Talati'', ``Patel'', ``Karnam'', ``Adhikari'', ``Shanbogaru'',``Patnaik'' etc.) is an administrative government position found in rural parts of the Indian sub-continent. The office and the officeholder are called the ``patwari'' in Telangana, Bengal, North India and in Pakistan while in Sindh it is called ``tapedar''. The position is known as the ``karnam'' in Andhra Pradesh, ``patnaik'' in Orissa or ``adhikari'' in Tamil Nadu, while it is commonly known as the ``talati'' in Karnataka, Gujarat and Maharashtra. The position was known as the ``kulkarni'' in Northern Karnataka and Maharashtra. The position was known as the ``shanbogaru'' in South Karnataka.
\\
Charles Craft (May 9, 1902 – September 19, 1968) was an English-born American film and television editor. Born in the county of Hampshire in England on May 9, 1902, Craft would enter the film industry in Hollywood in 1927. The first film he edited was the Universal Pictures silent film, ``Painting the Town''. Over the next 25 years, Craft would edit 90 feature-length films. In the early 1950s he would switch his focus to the small screen, his first show being ``Racket Squad'', from 1951–53, for which he was the main editor, editing 93 of the 98 episodes. He would work on several other series during the 1950s, including ``Meet Corliss Archer'' (1954), ``Science Fiction Theatre'' (1955–56), and ``Highway Patrol'' (1955–57). In the late 1950s and early 1960s he was one of the main editors on ``Sea Hunt'', starring Lloyd Bridges, editing over half of the episodes. His final film work would be editing ``Flipper\'s New Adventure'' (1964, the sequel to 1963\'s ``Flipper''. When the film was made into a television series, Craft would begin the editing duties on that show, editing the first 28 episodes before he retired in 1966. Craft died on September 19, 1968 in Los Angeles, California.
\\ \hline
\textbf{Query:}     What government position was held by the woman who portrayed Corliss Archer in the film Kiss and Tell?      \\ \hline
\textbf{Ground Truth:}   Shirley Temple Black (April 23, 1928 – February 10, 2014) was an American actress, singer, dancer, businesswoman, and diplomat who was Hollywood's number one box-office draw as a child actress from 1935 to 1938. As an adult, she was named United States ambassador to Ghana and to Czechoslovakia and also served as Chief of Protocol of the United States. Kiss and Tell is a 1945 American comedy film starring then 17-year-old Shirley Temple as Corliss Archer.   \\ \hline
\textbf{CSA Summary:}   Shirley Temple Black (April 23, 1928 – February 10, 2014) was an American actress, singer, dancer, businesswoman, and diplomat who was Hollywood's number one box-office draw as a child actress from 1935 to 1938. A Kiss for Corliss is a 1949 American comedy film directed by Richard Wallace and written by Howard Dimsdale.   \\ 
\hline
% \end{tabular}
\end{longtable}

\newpage
\subsection{Debatepedia}

In the visualization of the cross-attention score $p_i$, we use the following stop words to avoid messiness: ``to'', ``is'', ``and'', ``in'', ``the'', ``of''. For all visualizations, the darker color represents the larger value.

\begin{figure}[h]
% \vspace{-10pt}
\begin{center}
%\framebox[4.0in]{$\;$}
\includegraphics[width=0.6\textwidth]{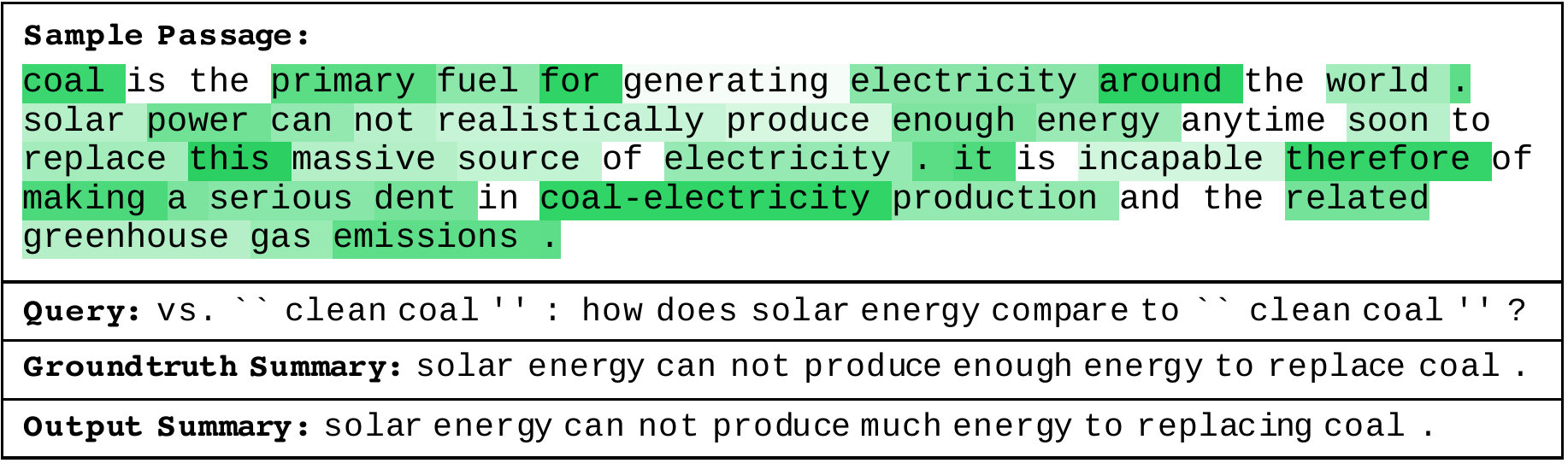}
\end{center}
\vspace{-10pt}
\caption{\label{fig:p1} Sample passages with attention. Highlighted with $p_i$.}
\end{figure}

\begin{figure}[h]
% \vspace{-10pt}
\begin{center}
%\framebox[4.0in]{$\;$}
\includegraphics[width=0.9\textwidth]{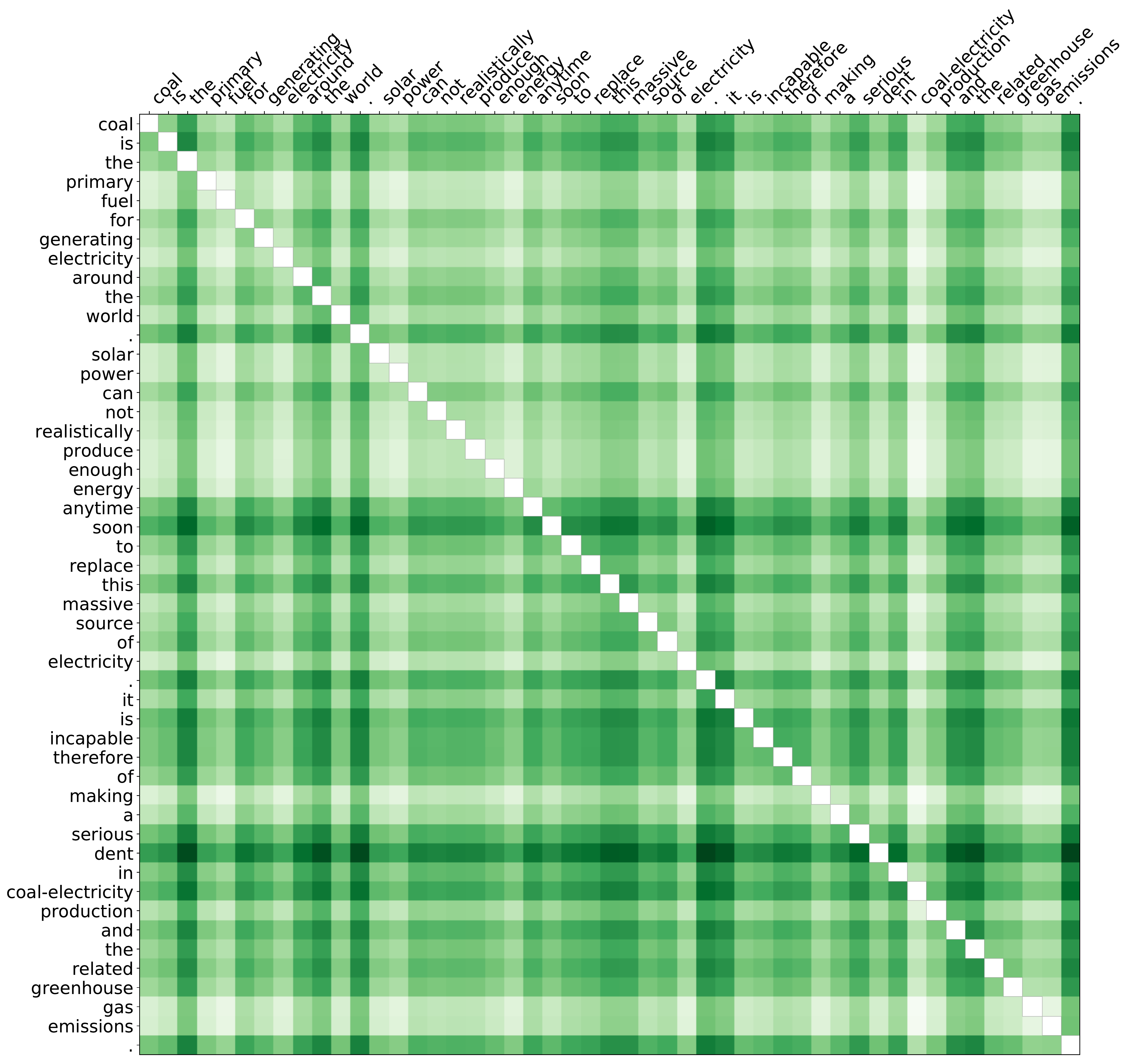} 
\end{center}
\vspace{-10pt}
\caption{Sample attention score of $f_{\text{csa}}(h_i,h_j)$ corresponding to the passage in Figure \ref{fig:p1}.}
\end{figure}

\newpage
\begin{figure}[h]
\vspace{-10pt}
\begin{center}
%\framebox[4.0in]{$\;$}
\includegraphics[width=0.7\textwidth]{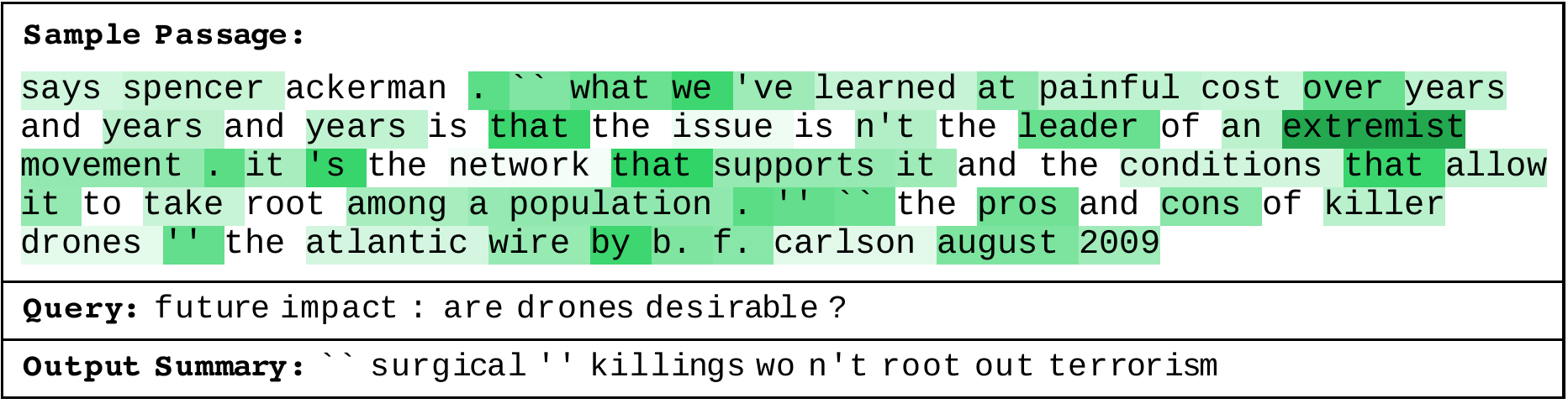}
\end{center}
\vspace{-10pt}
\caption{\label{fig:p2} Sample passages with attention. Highlighted with $p_i$.}
\end{figure}

\begin{figure}[h]
\vspace{-10pt}
\begin{center}
%\framebox[4.0in]{$\;$}
\includegraphics[width=0.9\textwidth]{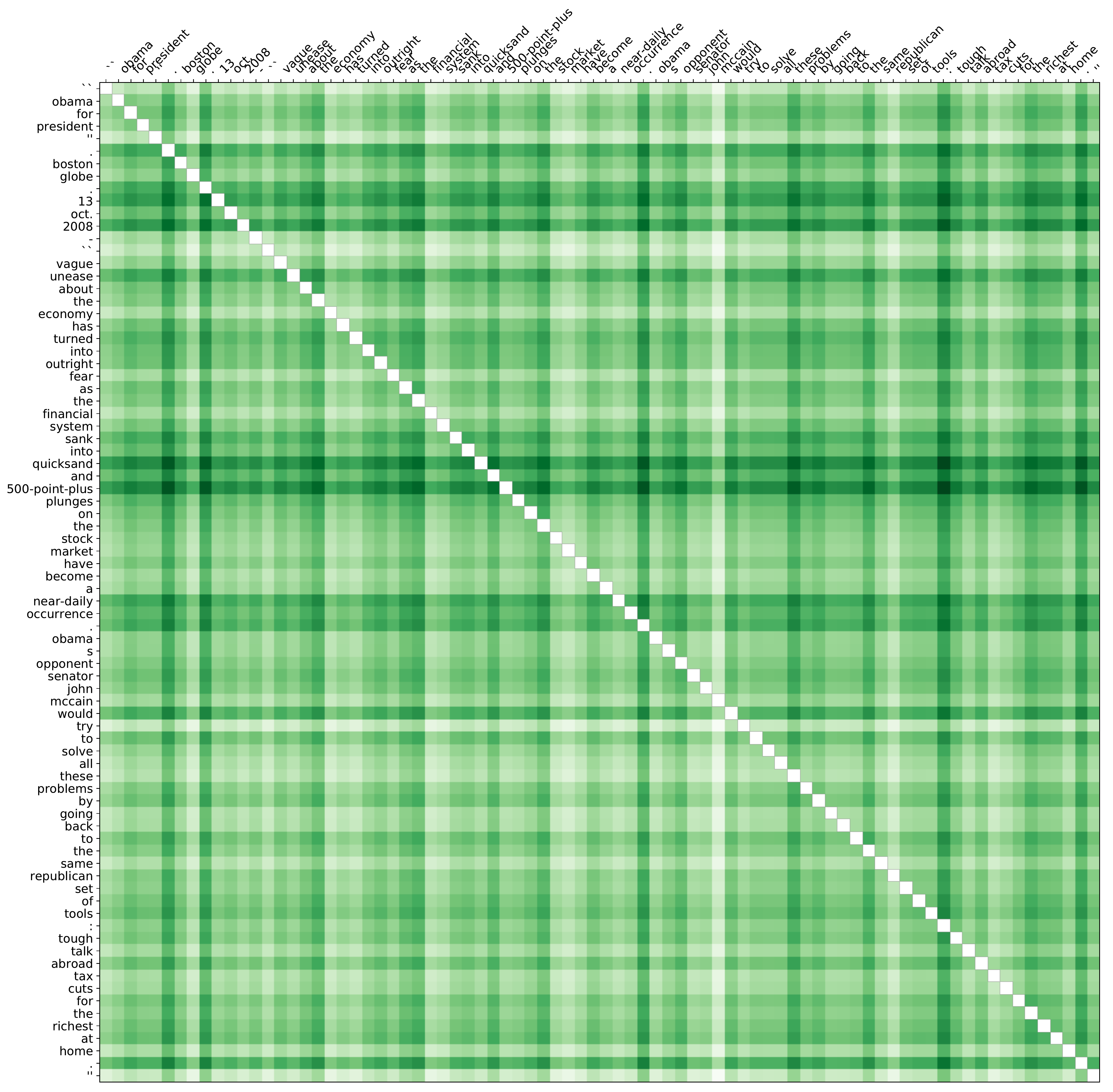} 
\end{center}
\vspace{-10pt}
\caption{Sample attention score of $f_{\text{csa}}(h_i,h_j)$ corresponding to the passage in Figure \ref{fig:p2}.}
\end{figure}

\newpage

\begin{figure}[t]
% \vspace{-10pt}
\begin{center}
%\framebox[4.0in]{$\;$}
\includegraphics[width=0.7\textwidth]{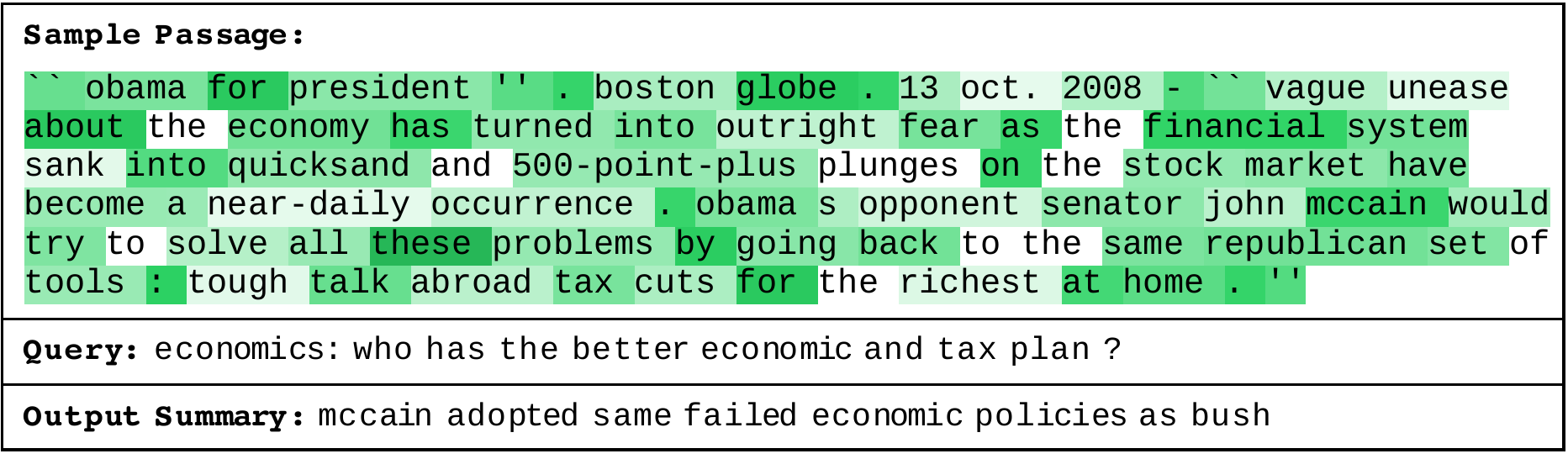} 
\end{center}
\vspace{-10pt}
\caption{\label{fig:p3} Sample passages with attention. Highlighted with $p_i$.}
\end{figure}

\begin{figure}[t]
% \vspace{-10pt}
\begin{center}
%\framebox[4.0in]{$\;$}
\includegraphics[width=0.98\textwidth]{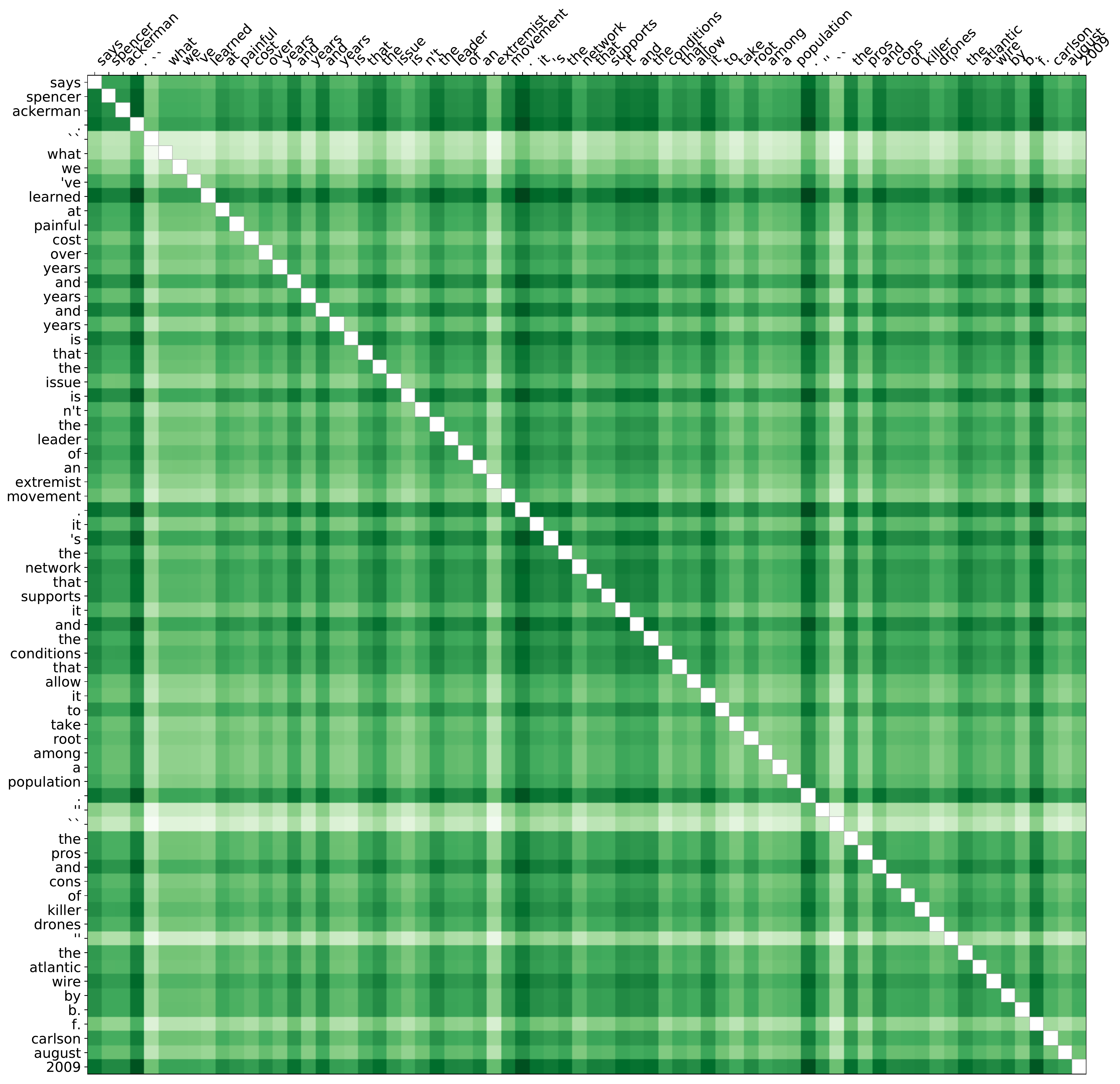} 
\end{center}
\vspace{-10pt}
\caption{Sample attention score of $f_{\text{csa}}(h_i,h_j)$ corresponding to the passage in Figure \ref{fig:p3}.}
\end{figure}

\end{document}